\documentclass[10pt]{article}
\usepackage{algorithm}
\usepackage{algorithmic}
\usepackage[preprint]{tmlr} 

\usepackage[utf8]{inputenc}
\usepackage{microtype}
\usepackage{wrapfig}
\usepackage{graphicx}
\usepackage{amsmath,amssymb}
\usepackage{subfigure}
\usepackage{booktabs}
\usepackage{multirow}
\usepackage[table]{xcolor}
\usepackage[skins]{tcolorbox}
\usepackage{adjustbox}
\usepackage{xspace}
\usepackage{url}
\usepackage{hyperref}
\hypersetup{hidelinks}

\newcommand{\algname}{\textsc{ACSeL}\xspace}

\title{\algname: \underline{A}daptive \underline{C}onfiguration \underline{Se}arch for Hardware-Constrained \underline{L}LM Inference}

\author{\name Bokun Wang, Hung-Yueh Chiang, and Diana Marculescu \\
	\addr University of Texas at Austin}
\date{}

\begin{document}
	
	\maketitle
	
	\let\AND\relax
	
	\begin{abstract}
		Serving large language models (LLMs) under tight memory and latency service-level objectives
		(SLOs), including on edge devices, requires jointly selecting the model size, batch size,
		quantization algorithm, and mixed-precision quantization recipes. Exhaustive search over
		this combinatorially large space is infeasible. Unlike data-center serving, which often
		prioritizes throughput or generation speed, resource-constrained deployments typically set
		memory and latency budgets as constraints and then seek the configuration with the highest
		generation quality among feasible ones. To this end, we propose ACSeL, a framework
		that identifies configurations predicted to satisfy deployment constraints while achieving
		high predicted generation quality. To enable efficient quality estimation, we use
		Relative Model Quality Degradation (RMQD), a proxy based on the divergence between
		the candidate and reference models' next-token distributions over a small calibration corpus.
		Across the evaluated model sizes and precisions, RMQD is inexpensive to evaluate and
		strongly correlated with downstream benchmark performance. ACSeL combines
		(i) a lightweight quality surrogate that predicts RMQD directly from a configuration;
		(ii) Few-shot Median Calibration (FMC), which calibrates operator-profile-based latency
		estimates using a small number of end-to-end profiles; and (iii) a structured search algorithm BandPareto
		for exploring layerwise mixed-precision configurations. Experiments on an NVIDIA RTX A6000 workstation and an NVIDIA Jetson Thor edge platform show that FMC substantially improves deployment feasibility, while BandPareto + FMC achieves the highest feasibility and the lowest RMQD gaps to the best measured configurations in the evaluation set.
	\end{abstract}
	
	\section{Introduction}\label{sec:intro}
	
	Large language models (LLMs) have advanced rapidly and are increasingly deployed beyond the datacenter, on edge devices, personal workstations, vehicles, wearable patient monitors, and interactive appliances that need on-device or near-device reasoning with low latency and strong privacy. These platforms face strict \emph{physical limits}: finite DRAM and flash storage, limited accelerator memory bandwidth, relatively strict interactive latency budgets such as time-to-first-token (TTFT) and time-per-output-token (TPOT) that users experience as responsiveness, and hard thermal and power constraints. Under such complex conditions, maximizing generation quality usually means optimizing over a joint \emph{configuration space} for a given model family and serving backend: the product of global choices (such as model size and batch size) and a per-layer quantization configuration space, where each layer specifies weight/activation formats, quantization algorithm, quantization scope (\emph{e.g.}, full or Linear-only), and options grouping layers into bands that share a recipe across contiguous layers. Exploring such a complex space is challenging because it is combinatorially large: for example, the number of candidate configurations is already on the order of $10^7$ for Qwen3 family models and Ampere-class GPUs, rendering any exhaustive profiling impractical. This motivates an adaptive configuration optimization problem: for each deployment request, automatically select the configuration that maximizes generation quality subject to its hardware and latency budgets.
	
	Prior work on configuration optimization for LLM serving typically targets data-center objectives, such as throughput or energy under latency SLOs, using control knobs such as parallelism and runtime scheduling~\citep{xu2026aiconfigurator,chooptimizing,song2026energylens}. These methods also generally assume fixed and  uniform quantization recipes. On hardware-constrained platforms the picture changes. Deployments often run on a single accelerator at modest request rates, so multi-GPU parallelism and cluster-scale scheduling offer little headroom, while latency budgets act as SLOs rather than soft targets to trade against tokens/sec. For example, a wearable patient monitor cannot add another rack, and the operator seeks to ensure that each response satisfies the latency constraints while achieving the highest possible generation quality. Layerwise weight/activation precision provides another lever for meeting memory and latency constraints while preserving generation quality. We therefore consider the complementary objective of maximizing the generation quality subject to latency and memory constraints:
	\begin{align}\label{eq:opt}
		&\max_{C\in \mathcal{C}_q}  F(C)
		,\quad \text{s.t.} \quad   \text{TTFT}(C,q) \leq L_1, ~\text{TPOT}(C,q) \leq L_2, ~ \text{Mem}(C,q) \leq m_0,
	\end{align}
	where $q=$\{hardware, \textit{isl}, \textit{osl}, $L_1$, $L_2$, \text{model family}, \text{backend}\} is a deployment request, \textit{isl}, \textit{osl} refer to input/output sequence lengths, $L_1$, $L_2$ are constraints on TTFT/TPOT, $C$ is a configuration from space $\mathcal{C}_q$ for request $q$, $F(C)$ measures model quality (instantiated later via an inexpensive proxy), and $m_0$ is the memory constraint of the given hardware platform\footnote{\algname could also incorporate energy constraints conservatively as latency $\times$ thermal design power limit.}.
	
	\begin{figure}[t]
		\centering
		\includegraphics[width=0.65\linewidth]{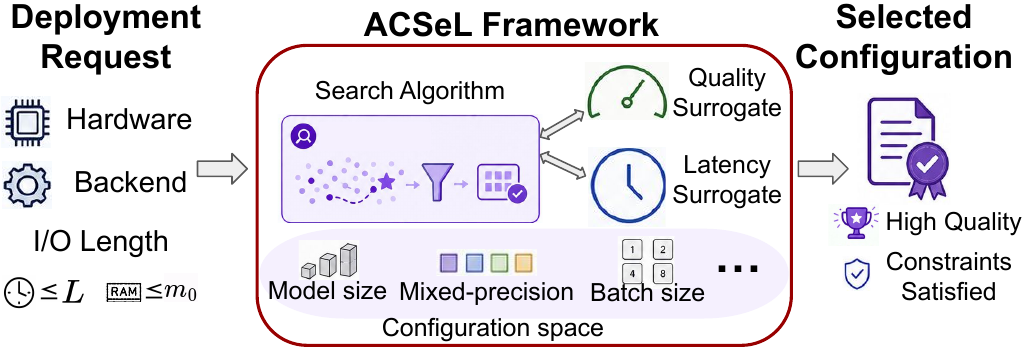}
		\caption{Overview of \algname, which uses quality and latency surrogates to guide the search over the configuration space and returns the predicted-feasible configuration with the highest predicted generation quality for the given deployment request.}
		\label{fig:framework}
	\end{figure}
	
	To efficiently solve~\eqref{eq:opt}, we present \algname (illustrated in Figure~\ref{fig:framework}), a surrogate-guided framework for jointly selecting the model size, serving configuration, and layerwise mixed-precision quantization recipe under request-specific constraints. \algname replaces costly benchmark evaluation and exhaustive hardware profiling with lightweight quality and latency surrogates, and then uses these surrogates to search the combinatorially large configuration space for a  feasible solution with highest predicted generation quality. Our main contributions are as follows:
	
	$\bullet$  We use \emph{Relative Model Quality Degradation} (RMQD), defined as the KL divergence between the next-token distributions of candidate models and a shared reference model over a small calibration corpus, as a low-cost quality proxy. We further develop a lightweight surrogate that predicts RMQD directly from configurations.
	
	$\bullet$ We develop \emph{Few-shot Median Calibration} (FMC), which improves AIConfigurator~\citep{xu2026aiconfigurator} using only a small number of end-to-end profiles and robust log-space median regression, improving measured deployment feasibility with limited platform-specific profiling.
	
	$\bullet$ We design efficient surrogate-guided search algorithms, including our \emph{BandPareto (BP)}, to enable efficient exploration of combinatorial mixed-precision configuration spaces under deployment-time constraints. Experiments on a workstation with an NVIDIA RTX A6000 GPU and an NVIDIA Jetson Thor edge platform show that FMC substantially improves deployment feasibility, while BandPareto + FMC achieves the lowest RMQD gaps. Our collected latency dataset of mixed-precision quantized LLMs could aid future research in resource-constrained serving.
	

	\section{Related Work}

	\paragraph{Post-Training Quantization (PTQ) for LLMs.} Numerous studies reduce the computational and memory costs of LLM inference through weight-only/weight-activation quantization~\citep{frantar2022gptq,dettmers2022gpt3,lin2024awq,xiao2023smoothquant}. 
	Another line of work explores mixed-precision PTQ, retaining sensitive components at higher precision while quantizing the remainder more aggressively~\citep{lee2024owq,saxena2025resq,zhao2026coopq}. These methods improve quality-efficiency trade-offs but typically do not adapt quantization to request-specific SLOs. AMQ~\citep{lee2025amq} searches layerwise bit-width assignments for weight-only quantization under a memory budget. In contrast, our work considers a broader deployment configuration space, including model and batch sizes as well as quantization recipes beyond bit width. Moreover, ACSeL explicitly models hardware- and backend-specific inference latency to optimize configurations under request-specific SLOs.
	

	\paragraph{Performance Modeling of LLM Inference.}
	
	Direct profiling provides accurate platform-specific latency measurements and bottleneck diagnosis~\citep{you2023zeus,chiang2025elana,wang2025lm}. However,  profiling every combination of possible configurations remains prohibitively expensive. 
	A growing body of work aims to predict the latency of LLM inference without exhaustively measuring every hardware and configuration point. 
	\citet{imai2024predicting} train regression models on profiled latency data with roofline-derived analytical latency~\citep{yuan2024llm} as an additional feature. NeuSight~\citep{lee2025forecasting} decomposes a kernel into tiles, uses an MLP to predict per-tile device utilization, and bounds the result with fundamental roofline performance laws. 
	AIConfigurator (AIC)~\citep{xu2026aiconfigurator} estimates TTFT/TPOT by offline operator profiling and table lookup: For each backend it builds a per-operator database keyed by type and shape, decomposes a prefill or decode step into operators, and sums their latencies, enabling latency prediction on GPUs without end-to-end latency profiling. We find that using raw AIC as the latency surrogate in configuration optimization yields low deployment feasibility under TTFT/TPOT SLOs. To this end, we propose a lightweight few-shot calibration of AIC that substantially improves feasibility with a small overhead.

	\paragraph{Deployment Optimization for LLM Serving.}
	
	Modern LLM serving systems such as vLLM~\citep{kwon2023efficient}, SGLang~\citep{zheng2024sglang}, and TensorRT-LLM~\citep{nvidia-trtllm} provide high-throughput engines with continuous batching, paged KV caches, and backend-specific kernels. On top of these systems, prior work considers the configuration optimization problem with data-center centric objectives and control knobs:  AIC~\citep{xu2026aiconfigurator} searches parallelism strategies (tensor/pipeline/expert) and framework runtime parameters (\emph{e.g.}, CUDA graphs, KV-cache memory fraction) across TRT-LLM, vLLM, and SGLang to improve serving performance under latency/throughput targets. 
	\citet{chooptimizing} analyze large deployment spaces spanning hardware choice, parallelism, and runtime configurations 
	aiming to maximize throughput subject to latency SLOs.
	EnergyLens~\citep{song2026energylens} explores multi-GPU parallelism, fusion, and compute-communication overlap to reduce inference energy. To accommodate the hardware-constrained LLM inference, our setting instead optimizes generation quality on single-accelerator, hardware-constrained platforms subject to memory and request-specific latency constraints, with (layerwise) quantization recipes in the search space.

	\section{Method}
	
	The optimization problem in \eqref{eq:opt} cannot be solved directly at deployment time: Evaluating model quality on standard benchmarks and profiling latency on the target hardware for every candidate configuration $C\in\mathcal{C}_q$ would incur prohibitive computational and runtime costs. To this end, we first define the Relative Model Quality Degradation (RMQD) score, an inexpensive and continuous metric that measures the quality degradation of a configuration relative to a family-specific oracle to replace direct benchmark evaluation (Section~\ref{sec:rmqd}). We then instantiate the quality objective as $F(C)=-\text{RMQD}(C)$, so maximizing quality in \eqref{eq:opt} becomes minimizing RMQD.  
	Because directly computing RMQD and measuring latency for every candidate configuration is still expensive, we approximate them with learned surrogates and solve:
	\begin{align}\label{eq:opt_emp} &\min_{C\in \mathcal{C}_q} \widehat{\text{RMQD}}(C) ,\\\nonumber & \text{s.t.} \quad \widehat{\text{TTFT}}(C,q) \leq L_1, ~\widehat{\text{TPOT}}(C,q) \leq L_2, ~ \text{Mem}(C,q) \leq m_0, \end{align}
	where $\widehat{\text{RMQD}}(C)$ denotes the predicted RMQD of configuration $C$, while $\widehat{\mathrm{TTFT}}(C,q)$ and $\widehat{\mathrm{TPOT}}(C,q)$ denote its predicted TTFT and TPOT, respectively, given deployment request $q$. Our framework \algname consists of three components: (i) lightweight quality surrogate that predicts $\widehat{\text{RMQD}}(C)$ from the configuration without requiring benchmark evaluation; (ii) latency surrogate
	$\widehat{T}(C,q)=\bigl(\widehat{\mathrm{TTFT}}(C,q),\widehat{\mathrm{TPOT}}(C,q)\bigr)$, that extends the operator-level lookup table of AIC to device-specific kernels and layerwise quantization recipes and calibrates its predictions using few-shot profiling and median regression; and (iii) search algorithms that use the quality and latency surrogates to solve \eqref{eq:opt_emp} efficiently.
	
	\subsection{Relative Model Quality Degradation (RMQD)}\label{sec:rmqd}
	
	Two natural model quality metrics are test loss (\emph{e.g.}, cross-entropy loss on a held-out corpus) and Benchmark scores (\emph{e.g.}, on NeMo Evaluator~\citep{nvidia-nemo}). Test loss is smooth and therefore convenient for fitting
	predictors. However, it is an absolute metric: a practitioner cannot read off ``how good'' a quantized 8B model is from
	a raw loss number without a reference. Moreover, corpus loss might not reliably reflect common-sense, mathematical, or
	code reasoning quality under deployment-relevant prompts. On the other hand, benchmark scores are directly interpretable to practitioners, but they are
	essentially step functions of the underlying logits, which are ill-suited as regression targets for a small-data ML surrogate. Furthermore, evaluating every candidate configuration on standard benchmarks can be time-consuming and expensive in terms of token usage and API costs. 
	
	
	
\begin{figure}[h]
		\subfigure{
			\centering
			\includegraphics[width=0.325\linewidth]{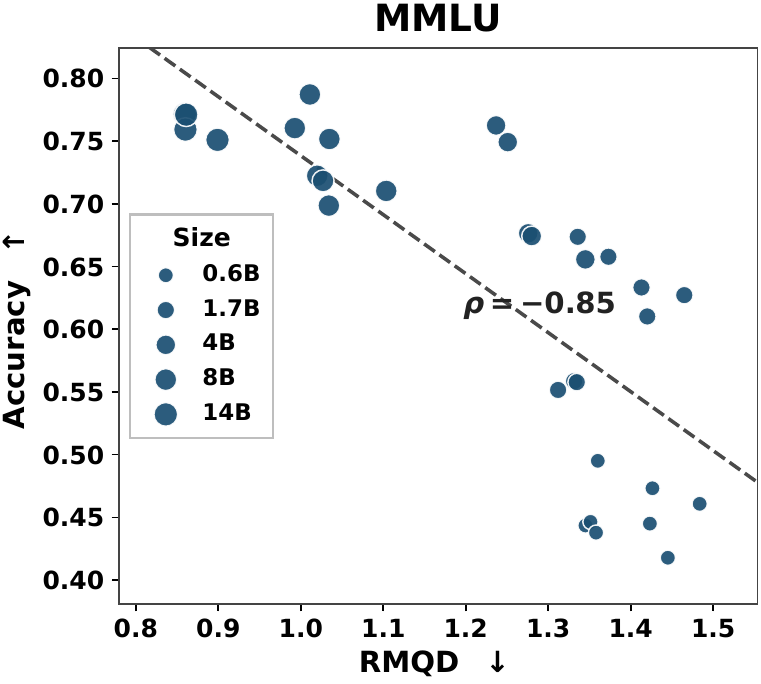}
		}
		\subfigure{	\centering
			\includegraphics[width=0.325\linewidth]{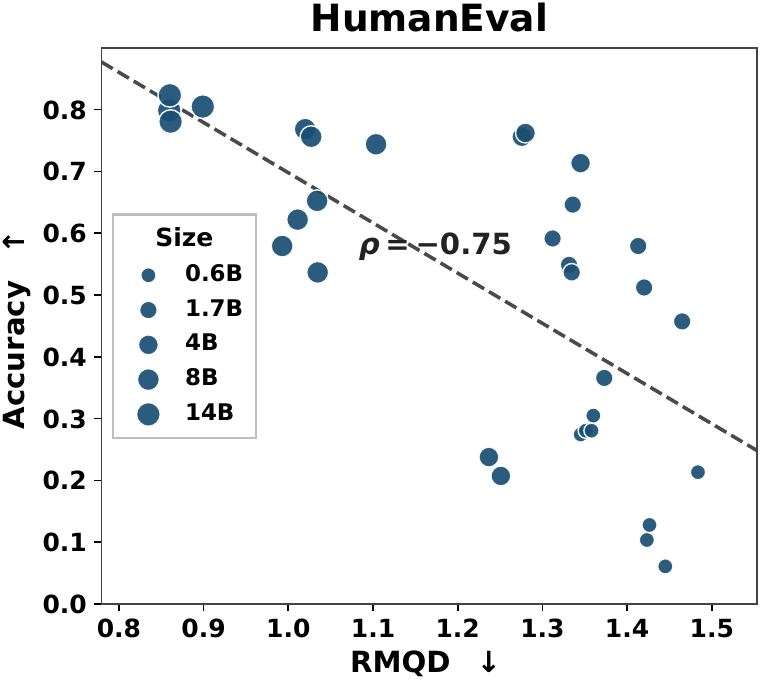}
		}
		\subfigure{
			\centering
			\includegraphics[width=0.325\linewidth]{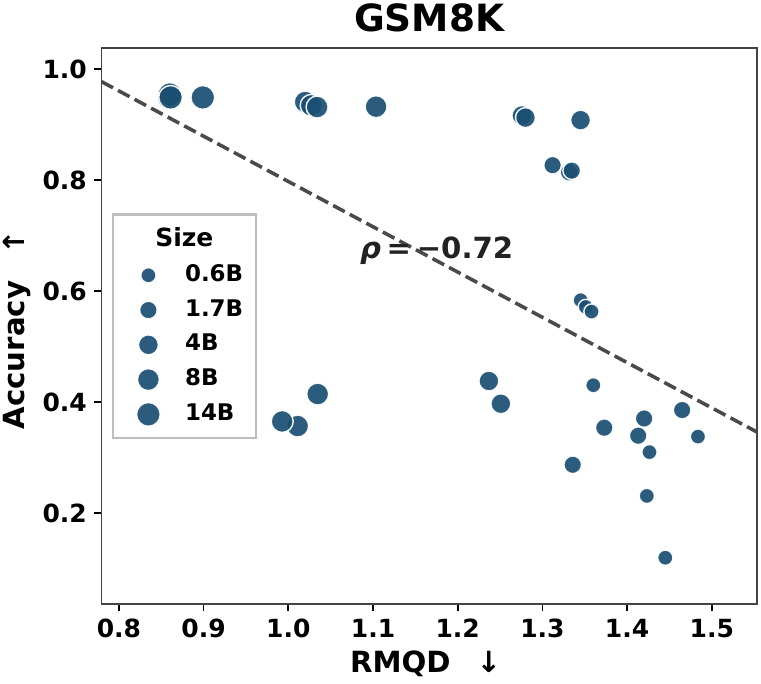}
		}
		\caption{Spearman correlation ($\rho$) between accuracy and RMQD for Qwen3 models with both uniform-precision and mixed-precision quantization recipes across three benchmarks from NeMo Evaluator. RMQD is fairly strongly negatively correlated with accuracy on all benchmarks ($\rho=-0.72$ to $-0.85$), capturing the overall
			quality trend across model sizes and precisions. Dashed lines show linear fits.}
		\label{fig:kl_vs_acc}
	\end{figure}

	To obtain a quality metric that is \emph{continuous}, \emph{more interpretable} than raw test loss, and \emph{cheap} to evaluate, KL divergence between a compressed model and a high-precision reference has been used as an inexpensive fidelity metric for compressed LLMs, complementing benchmark accuracy \citep{dutta2024accuracy, nikolic2026displacement}. We build on this idea but adapt it to ACSeL's cross-model configuration search. In contrast to comparing each quantized model only with its own full-precision counterpart, we use a common family-level oracle, which places different model sizes and quantization recipes on a shared reference scale. We call the resulting corpus-averaged score \emph{Relative Model Quality Degradation} (RMQD): 
	Consider a model family $\mathcal{F}$ with a shared tokenizer and vocabulary $\mathcal{V}$ (\emph{e.g.}, Llama-3). Let the oracle $M^*\in \mathcal{F}$ be the
	largest full-precision model in that family (\emph{e.g.}, Llama-3.1 405B). For a candidate
	configuration $C$ inducing model $M_C$ (same tokenizer/vocabulary), and a sequence $x = (x_1,\dotsc, x_L)\in \mathcal{V}^L$
	, denoted by $p^*(\cdot\mid x_{}<t)$ and $p_C(\cdot\mid x_{}<t)$  the oracle and candidate next-token distributions at position $t$. The sequence-level RMQD is
	\begin{align*}
		\text{RMQD}(C; x)=\frac{1}{L} \sum_{t=1}^L D_{\mathrm{KL}}\left(p^{\star}\left(\cdot \mid x_{<t}\right) \|p_C\left(\cdot \mid x_{<t}\right) \right),
	\end{align*}
	where $D_{\mathrm{KL}}$ is the KL-divergence, given a fixed calibration corpus $\mathcal{D}$ of sequences subsampled from diverse benchmarks. In our work, we use 24 samples from each of the 17 benchmarks of NeMo Evaluator. The overall RMQD score is $\text{RMQD}(C) = \frac{1}{|\mathcal{D}|}\sum_{x\in \mathcal{D}} \text{RMQD}(C; x)$. A lower RMQD score indicates less quality degradation. As illustrated in Figure~\ref{fig:kl_vs_acc}, RMQD exhibits a fairly strong negative
	correlation with benchmark scores across the evaluated model sizes
	and quantization recipes. These results support RMQD as a computationally efficient
	proxy for the overall quality trend.
	
	\paragraph{Computational efficiency.} Oracle inference dominates the cost which is hardware-agnostic and incurred only once: the oracle model's logits can be computed on the cloud/cluster, cached, and reused to evaluate any candidate model on the target platform. Because caching full-vocabulary logits would incur substantial storage and I/O overhead, we retain only the oracle's top-$m$ token probabilities ($m=512$ in our experiments, covering ~99\% of the probability mass on the oracle models we consider) and collapse the remaining mass into a single tail bin, yielding a $(m+1)$-bin distribution.
	
	Direct benchmark evaluation typically requires sequential autoregressive decoding to generate an answer token by token, followed by answer verifier, exact-match checker, LLM-as-a-judge, or external API calls. In contrast, computing RMQD requires only a single teacher-forced forward pass over the fixed calibration sequences and can be parallelized across both tokens and sequences. Empirically, on an A6000, computing RMQD for 24 GSM8K questions takes 6.84 seconds, compared with 626.91 seconds for directly evaluating benchmark scores on the same 24 questions (91.6$\times$ speedup).

	\subsection{Quality Surrogate}\label{sec:quality}
	
	Section~\ref{sec:rmqd} gives a cheap quality proxy, but computing RMQD score for each candidate LLM still requires running a teacher-forced forward pass. We therefore fit a lightweight linear model to map (layerwise) configuration $C$ to predicted $\widehat{\text{RMQD}}(C)$,  evaluated in milliseconds at search time.
	
	We preprocess each input configuration $C$ into numerical features $\phi(C)$, which is the concatenation of three groups of features as follows. First, we treat the logarithm of the model size and the one-hot encoding of the quantization algorithm as \emph{free} features. Second, we represent the exclusion flags (indicating whether the LM head, MoE gate, and KV cache are quantized) as \emph{global} features. Third, for each layer $\ell$, we encode the bit deficit, defined as $16-$ \#bits, the quantization scope (full layer, MLP only, \emph{etc.}), and the quantization granularity (per-tensor, per-channel, block-128, or block-16) as \emph{numerical} features normalized to $[0,1]$. Then, we aggregate the layerwise features into global features by both uniform averaging and depth-weighted averaging which puts weights on early, middle, and late layers. We also impose the prior that the model degradation is monotonic in bit deficit features (Ablation study of monotone features is in Section~\ref{sec:ablation}).  
	
	In practice, the measured RMQD scores span multiple orders of
	magnitude and are strictly positive. We therefore model the target $y(C)=\log(\text{RMQD}(C))$ using a linear model on
	$\phi(C)$ and fit the quality surrogate by minimizing a constrained
	ridge-regression objective:
	\begin{align*}
		& \min _{w, b} \sum_i(w^{\top} \phi(C^{(i)})+b-y^{(i)})^2+\lambda\|(w,b)\|_2^2,\\
		& \mathrm{s.t.}~w_j\geq 0,\forall j\in M,
	\end{align*}
	where $M$ contains the coordinates of all monotone features. 
	The quality surrogate is fit once offline on a small set of configurations. Since the RMQD score depends only on model and quantization configurations, the cost of fitting can be  amortized across model sizes, hardware platforms, backends, and subsequent deployment requests. Search-time RMQD prediction is
	therefore a milliseconds-scale feature map and dot product, with no additional oracle or benchmark evaluation in the
	inner loop.

	\subsection{Latency Surrogate}\label{sec:latency_surrogate}
	
	AIConfigurator (AIC)~\citep{xu2026aiconfigurator} decomposes prefill/decode into operators and estimates latency by table lookup over
	profiled kernels. We build on AIC but extend it to
	our deployment setting in three steps. First, we extend the operator-level lookup tables to target hardware platforms A6000 and Jetson Thor. Besides, vanilla AIC targets uniform numerical formats. For a multi-band mixed-precision quantization configuration $C=\{(C_b, \mathcal{I}_b)\}_{b=1}^B$, where $\mathcal{I}_b$ is the layer index set of the $b$-th band, we first obtain per-band uniform AIC estimates for each band and integrate them into end-to-end latency prediction. 
	
	Even after the aforementioned extensions, directly using AIC predictions to filter candidates results in a low feasibility rate: many configurations predicted to satisfy the TTFT or TPOT SLOs violate them when deployed on the target hardware (See experimental results in Section~\ref{sec:ablation}). To improve prediction with minimal profiling overhead, we propose the Few-shot Median Calibration (FMC). First, we select $K$ configurations per backend and phase (prefilling or decoding) by spreading them out across AIC-predicted latencies and recipe types. Starting from an initial configuration, we repeatedly select the candidate whose distance to its nearest previously selected configuration is largest. For each selected configuration $C^{(k)}$, ($k=1,\dots,K$), we record its measured latency $T^{(k)}$ and corresponding AIC prediction $\widehat{T}_{\mathrm{AIC}}^{(k)}$. For each phase, FMC models the calibration as
	\begin{align*}
		\log T = \gamma^* \log \widehat{T}_{\text{AIC}} + \delta^*, 
	\end{align*}
	where $(\gamma^*,\delta^*)$ is estimated by minimizing the pinball loss~\citep{koenker2001quantile} in the log-space:
	\begin{align}\label{eq:pinball}
		\min _{\gamma, \delta} \sum_{k=1}^K \rho_{0.5}(\log T^{(k)}-\gamma \log \widehat{T}_{\mathrm{AIC}}^{(k)}-\delta),
	\end{align}
	where $\rho_\tau(u) = u(\tau - \mathbf{1}_{u<0})$. Then, the predicted latency is  $\widehat{T}(C,q) = \exp(\gamma^* \log \widehat{T}_{\mathrm{AIC}}(C,q) + \delta^*)$ given configuration $C$ and request $q$. With  $\tau=0.5$, minimizing the pinball loss reduces to median regression, which is less sensitive than least-squares regression to large response residuals, which is particularly useful when only a small number of profiled data is available (see Section~\ref{sec:ablation} for an empirical comparison). We fit the calibration model in log space because AIC errors are primarily multiplicative. For example, predicting $10$ ms instead of $20$ ms and $200$ ms instead of $400$ ms represents the same $2\times$ underestimation. Log-space regression assigns both cases the same residual, $\log 2$, preventing large-latency configurations from dominating the regression objective. 
	
	\paragraph{Computational efficiency.} FMC requires only a one-time, offline campaign of few (tens of) end-to-end latency profiles per serving stack (device, backend) and solving a milliseconds-scale univariate median regression. The fitted surrogate can be reused for every model, configuration, and deployment requests.

	\subsection{Search Algorithms}
	
	Given the calibrated latency surrogate $\mathrm{FMC}(\widehat{T}_{\text{AIC}}(C,q))$, we construct a surrogate admission gate $\mathrm{Gate}(C,q)$ that determines whether a candidate configuration $C$ is predicted to satisfy the deployment constraints of request $q$. As shown in Algorithm~\ref{alg:eval}, the gate first rejects configurations that exceed the available memory or cannot be exported to the target backend. It then applies FMC to predict TTFT and TPOT and rejects configurations that violate either latency SLO.
	
	\begin{algorithm}[h]
		\caption{\textbf{Surrogate admission gate $\mathrm{Gate}(C,q)$}}
		\label{alg:eval}
		\begin{algorithmic}[1]
			\REQUIRE Configuration $C$, deployment request $q$, TTFT/TPOT constraints $L_1/L_2$, memory constraint $m_0$ 
			\IF{memory $>m_0$ \textbf{or} non-exportable}
			\STATE \textbf{return} \texttt{Reject}
			\ENDIF
			\STATE $(\widehat{\mathrm{TTFT}}(C,q),\widehat{\mathrm{TPOT}}(C,q))\gets\mathrm{FMC}(\widehat{T}_{\text{AIC}}(C,q))$
			\IF{$\widehat{\mathrm{TTFT}}(C,q)>L_1$ \textbf{or} $\widehat{\mathrm{TPOT}}(C,q)>L_2$}
			\STATE \textbf{return} \texttt{Reject}
			\ENDIF
			\STATE \textbf{return} \texttt{Accept}
		\end{algorithmic}
	\end{algorithm}
	
	Apart from batch size and model size, we explore  mixed-precision configurations represented as \emph{band recipes}. A band recipe partitions the layers of an LLM into contiguous bands. Each band is assigned a quantization configuration including a data format, quantization algorithm, scope (which tensors in a layer are quantized), and quantization granularity (how scaling factors are shared). In our work, we consider at most three bands, each with a different quantization configuration. We also assume that the quantization algorithm is shared across all layers and the granularity is set to the default associated with each data format, such as block-16 for \texttt{nvfp4}. The search algorithms therefore primarily determine the band boundaries and the data format assigned to each band. We investigate three search algorithms from least to most structured: random sampling (Rand, Algorithm~\ref{alg:rand}), hill-climbing (HC, Algorithm~\ref{alg:hc}), and BandPareto (BP, Algorithm~\ref{alg:bp}).
	
	Algorithm~\ref{alg:rand} draws random band configurations, runs the full surrogate admission gate on each, and returns the admitted draw with the lowest predicted RMQD. Hill-climb (Algorithm~\ref{alg:hc}) instead improves a warm start by searching local neighbors (\emph{e.g.}, stepping a band’s format along \texttt{int4} $\leftrightarrow$ \texttt{int8} $\leftrightarrow$ \texttt{bf16}, or nudging a cut by one layer). HC could repair a bad initial configuration, but it still explores only a neighborhood of that start.
	
	\begin{algorithm}[h]
		\caption{Random sampling (Rand)}
		\label{alg:rand}
		\begin{algorithmic}[1]
			\REQUIRE Request $q$; time budget $\bar{t}$
			\STATE $\mathcal{F}\leftarrow\emptyset$
			\WHILE{elapsed time $<\bar{t}$}
			\STATE Draw a random configuration $C$
			\IF{$\mathrm{Gate}(C;q)== $ \texttt{Accept}}
			\STATE $\mathcal{F}\leftarrow\mathcal{F}\cup\{C\}$
			\ENDIF
			\ENDWHILE
			\STATE \textbf{return} $\arg\min_{C\in\mathcal{F}}\widehat{\text{RMQD}}(C)$ if $\mathcal{F}\neq \emptyset$, else unanswered
		\end{algorithmic}
	\end{algorithm}
	
	\begin{algorithm}[h]
		\caption{Hill-climbing (HC)}
		\label{alg:hc}
		\begin{algorithmic}[1]
			\REQUIRE Request $q$; time budget $\bar{t}$
			\STATE $J(C)\gets\widehat{\text{RMQD}}(C)$ if
			$\mathrm{Gate}(C,q)$ is  \texttt{Accept}, else $\infty$
			\STATE Randomly select a recipe $C$; $\mathit{best}\gets C$
			\WHILE{elapsed time $<\bar{t}$}
			\FORALL{neighbor $C'$ of $C$}
			\IF{$J(C')<J(C)$}
			\STATE $C\gets C'$; $\mathit{best}\gets
			\arg\min_{X\in\{\mathit{best},C\}}J(X)$; \textbf{break}
			\ENDIF
			\ENDFOR
			\STATE If no better neighbor exists, randomly restart $C$ and set
			$\mathit{best}\gets\arg\min_{X\in\{\mathit{best},C\}}J(X)$
			\ENDWHILE
			\STATE \textbf{return} $\mathit{best}$ if $J(\mathit{best})<\infty$,
			else unanswered
		\end{algorithmic}
	\end{algorithm}

	We notice that the dominant cost is the admission gate, which for multi-band candidates triggers a full AIC call (looking up the operator-level tables and accumulating). Utilizing this observation, BandPareto (Algorithm~\ref{alg:bp}) soft-screens many legal configurations with a cheap approximate latency check, retains only the competitive (Pareto) shortlist, and applies the expensive full admission gate to that shortlist alone.

	For the cheap latency check, we use a layer-fraction mix of precomputed uniform AIC predictions, which we call FracMix. Since the number of uniform quantization configurations is orders of magnitude smaller than the band ones, we run AIC on uniform configurations and store the end-to-end latency in a lookup table. At query time, the FracMix approximate latency of a $B$-band configuration $C = \{C_j\}_{j=1}^B$ is computed as $
	\mathrm{FMC}\left(\sum_{j=1}^B \frac{L_j}{L} \widehat{T}_{\text{AIC}}(C_j,q)\right)$, where $C_j$ is the configuration of the $j$-th band, $L$ is the number of total layers, $L_j$ is the number of layers in the $j$-th band, and $\widehat{T}_{\text{AIC}}(C_j,q)$ is the precomputed AIC estimate of \emph{whole-model} with uniform configuration $C_j$\footnote{If there is a miss, we can use the linear interpolation of entries.}.
	
	\begin{algorithm}[h]
		\caption{BandPareto (BP)}
		\label{alg:bp}
		\begin{algorithmic}[1]
			\REQUIRE Request $q$
			\STATE $\mathit{soft} \gets \emptyset$
			\STATE \texttt{\# 1. Fast soft screening}
			\FOR{each legal band configuration $C$}
			\STATE $(\widetilde{\mathrm{TTFT}}(C,q),\widetilde{\mathrm{TPOT}}(C,q))\gets\mathrm{FracMix}(C,q)$ 
			\IF{$\widetilde{\mathrm{TTFT}}(C,q)\leq (1+\eta) L_1$ and $\widetilde{\mathrm{TPOT}}(C,q)\leq (1+\eta) L_2$}
			\STATE $\mathit{soft} \gets \mathit{soft} \cup \{C\}$
			\ENDIF
			\ENDFOR
			\STATE \texttt{\# 2. Pareto filter}
			\STATE $\mathit{front} \leftarrow \{C\mid C\in \textit{soft},~\nexists\,C'\in\mathit{soft}:~C'\prec C\}$\footnotemark
			\STATE $\mathcal{F}\leftarrow\emptyset$
			\STATE \texttt{\# 3. Full gate for Pareto shortlist}
			\FOR{each $C$ in $\textit{front}$}
			\IF{$\text{Gate}(C,q)== $ \texttt{Accept}}
			\STATE $\mathcal{F} \gets \mathcal{F} \cup \{C\}$
			\ENDIF
			\ENDFOR
			\STATE \textbf{return} $\arg\min_{C\in\mathcal{F}}\widehat{\text{RMQD}}(C)$ if $\mathcal{F}\neq \emptyset$, else unanswered
		\end{algorithmic}
	\end{algorithm}
	\footnotetext{We call $C'\prec C$ if a configuration $C'$ is no worse than $C$ on
		$(\widehat{\text{RMQD}},\widetilde{\mathrm{TTFT}},\widetilde{\mathrm{TPOT}})$ and strictly better on at least one.}
	
	Empirically, on Jetson Thor, FracMix takes $\sim$10 s to build the lookup table and costs
	{$\sim$0.15\,ms} per candidate versus
	{$\sim$60\,ms} for a full admission gate
	({$\sim$400$\times$} slower). Note that the table is built once per serving stack and then reused across deployment queries, so its offline CPU cost is amortized over all subsequent online FracMix lookups. 
	Because FracMix is noisier than the full admission gate, we retain candidates whose predicted latency is up to $\eta$ above the latency constraints, reducing false rejection of near-boundary configurations before full gating. In our experiments, $\eta = 5\%$ is sufficient to match the feasibility performance of BandPareto in Table~\ref{tab:aic_vs_FMC}. Using a larger $\eta$ sends more candidates through full evaluation, which increases the running time (See Appendix~\ref{sec:ablation-eta}). 
	After soft screening, BP keeps only configurations that are on the empirical quality–latency Pareto frontier built from the soft candidates. Those survivors are fully gated, and among any that pass, we pick the one with the best predicted RMQD.

	\section{Experiments}\label{sec:exps}
	
	
	We aim to answer the following research question:
	
	\begin{tcolorbox}[
		colback=gray!8,
		colframe=gray!45,
		boxrule=0.7pt,
		arc=1.5mm,
		left=2.5mm,
		right=2.5mm,
		top=1.5mm,
		bottom=1.5mm
		]
		\emph{\textbf{RQ}: How effectively can surrogate-guided search identify deployment-feasible configurations with quality close to the best one under request-specific SLOs?}
	\end{tcolorbox}
	
	We also conduct ablation studies and comparisons with baseline methods to validate our design choices.

	\subsection{Experimental Setup}
	
	\paragraph{Platforms, backends, formats.} We consider two complementary deployment settings: an NVIDIA RTX A6000 workstation GPU and an NVIDIA Jetson Thor edge platform. On A6000 we cover the backends vLLM, SGLang, TensorRT-LLM, and data formats \texttt{bf16}, \texttt{int8}, \texttt{int4} (weight-only). On Thor, we cover the backends vLLM, SGLang\footnote{TensorRT-LLM is currently not supported on Thor~\citep{joostdev2025tensorrtllm}.}, and data formats \texttt{bf16}, \texttt{fp8}, \texttt{nvfp4} (weight-activation). 
	
	\paragraph{Evaluation data.} We consider candidate models from the Llama-3.x and Qwen3 families, with model sizes of up to 8B. On A6000, we profile 288 3-band and all uniform quantization configurations across batch sizes $\in\{1,4,8,16,32\}$, \textit{isl} $\in\{128,512,2048,8192\}$, and \textit{osl} $\in\{64,256,512\}$, yielding 13,618 latency measurements. On Thor, we profile 288 3-band and all uniform quantization configurations across batch sizes $\in\{1, 4, 8, 32\}$, \textit{isl} $\in\{128, 2048, 8192\}$, and \textit{osl} $\in\{64, 512\}$, yielding 4,309 latency measurements. For band quantization configurations, we profile the TTFT/TPOT on the target hardware after exporting the model using ModelOpt~\citep{nvidia-modelopt}. For the RMQD scores (which are hardware and request independent), we use Llama-3.1 405B and Qwen3 235B-A22B as the oracle models for
	the Llama-3.x and Qwen3 families, respectively. The oracle logits are computed and cached from a high-performance GPU cluster. 
	\begin{wrapfigure}{l}{0.6\textwidth}
		\subfigure{
			\centering
			\includegraphics[width=0.475\linewidth]{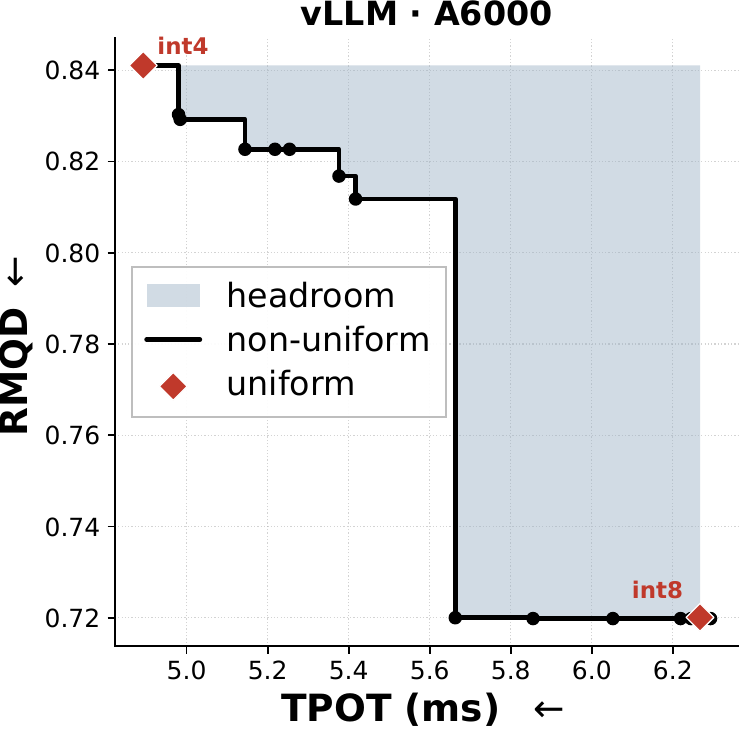}
		}
		\subfigure{	\centering
			\includegraphics[width=0.475\linewidth]{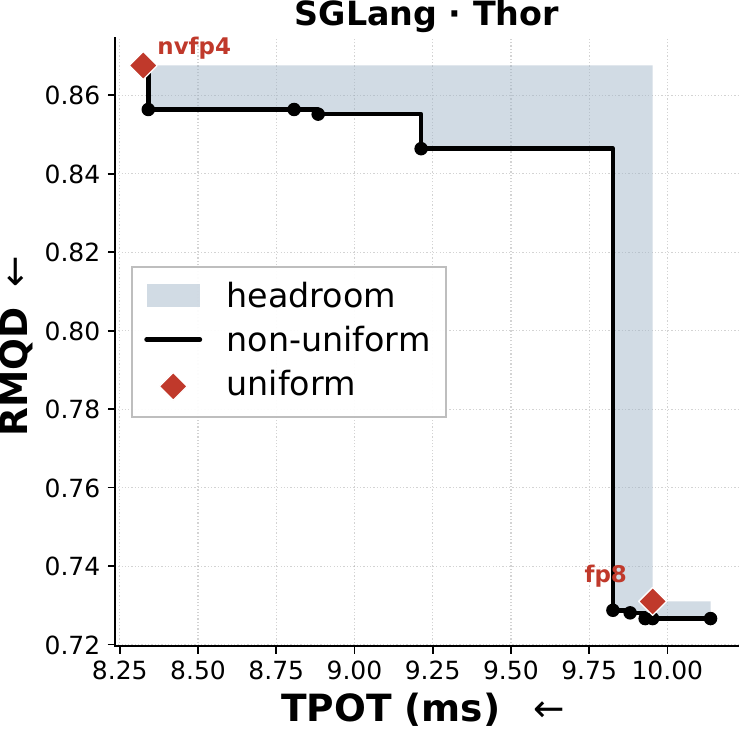}
		}
		\caption{Quality–latency headrooms on A6000 (vLLM) and Thor (SGLang) of band \emph{vs.} uniform in the evaluation dataset, with llama-3.2-1b at batch size 4, \textit{isl} 2048, \textit{osl} 64. 
		}
		\label{fig:headroom}
	\end{wrapfigure}
	For quantized candidate models, we apply ModelOpt's fake quantization. It is worth noting that none of the evaluation data is used to fit the quality or latency surrogates in our framework. Figure~\ref{fig:headroom} illustrates why mixed-precision configurations are useful under latency constraints. For a fixed model and workload, uniform quantization provides only a few quality–latency operating points. Band configurations offer additional choices between them: at intermediate TPOT budgets, they can achieve lower measured RMQD than the best measured uniform configuration satisfying the same budget. These examples motivate searching over band configurations rather than restricting deployment to uniform quantization.

	\subsection{Search Performance Comparison}\label{sec:performance}
	To compare search performance, we generate 50 synthetic deployment requests using a fixed procedure. Each request specifies a platform–backend pair and an $(\mathrm{isl},\mathrm{osl})$ workload. To construct the latency SLOs we pin an anchor model and batch size, then set the latency constraint to be tighter than the corresponding uniform int8 operating point on A6000 or uniform fp8 operating point on Thor. In this way we construct challenging but attainable latency SLOs in the mixed-precision headroom region illustrated in Figure~\ref{fig:headroom}, and the same request set is used
	for all algorithms (Full details are provided in Appendix~\ref{sec:requests}). For fair comparison, we set the time budget $\bar{t}$ in Rand and HC to match that of BP.
	We fit the quality surrogate described in Section~\ref{sec:quality} on 226 model configurations that are disjoint from the evaluation data. For variants with the FMC latency surrogate, we profile 30 configurations (evenly split across backends and disjoint from the evaluation data) according to the configuration selection approach described in Section~\ref{sec:latency_surrogate}.

	\paragraph{Metrics.} We report each search algorithm's \textbf{answer rate}: the fraction for which it returns a configuration that passes the surrogate admission gate. Then, we compare the feasibility rates of the configurations returned by the search algorithms. A configuration is deemed \textbf{feasible@tol\%} if it fits in memory, can be exported to the target backend, and its measured latency satisfies the deployment constraints on the target hardware within a tolerance of $\mathrm{tol}$\%, where $\mathrm{tol}\in\{0, 5,10\}$, where $\mathrm{tol}=0$ denotes strict feasibility.  To further measure how close the returned configurations are to the Pareto frontier, we also report \textbf{gap$\cap$@10\%}: the difference between the RMQD of the returned configuration and the best RMQD among all feasible configurations in the evaluation dataset, on requests where all compared algorithms are feasible@10\%. 
	We also report \textbf{$\geq$uniform@10\%}: the fraction of requests for
	which the returned configuration of a search algorithm matches or improves upon the RMQD
	of the best feasible@10\% uniform configuration. We run each algorithm with 3 random seeds and report the mean $\pm$ standard deviation.
	
	\begin{table}[h]
		\centering
		\caption{Answer rates, measured deployment feasibility, and quality
			of configurations returned by search algorithms with AIC and
			\textbf{FMC (Ours)}. 
			Feas.@0\% denotes strict satisfaction of the measured latency SLOs, whereas
			Feas.@5\% and Feas.@10\% allow the corresponding relative tolerance.
			$\mathrm{Gap}\cap@10\%$ is reported only for FMC variants and
			is computed (for each seed) over requests for which Rand, HC, and BP are all feasible@10\%. The reference is the best measured feasible@10\% configuration in the evaluation
			dataset.}
		\label{tab:aic_vs_FMC}
		\begin{adjustbox}{max width=\linewidth}
			\begin{tabular}{
					cc!{\vrule width 1.2pt}c|c|c|c
					!{\vrule width 1.2pt}c|c
				}
				\toprule[.1em]
				\multicolumn{8}{c}{\textbf{A6000}} \\
				\midrule
				\multicolumn{2}{c!{\vrule width 1.2pt}}{\textbf{Method}}
				& \textbf{Answered ($\uparrow$)}
				& \textbf{Feas.@0\% ($\uparrow$)}
				& \textbf{Feas.@5\% ($\uparrow$)}
				& \textbf{Feas.@10\% ($\uparrow$)}
				& \textbf{Gap$\cap$@10\% ($\downarrow$)}
				& \textbf{$\geq$uniform@10\% ($\uparrow$)} \\
				\midrule
				
				\multirow{2}{*}{Rand}
				& +AIC
				& 48.0\%$\pm$0.0\%
				& 30.7\%$\pm$6.1\%
				& 34.7\%$\pm$2.3\%
				& 40.0\%$\pm$4.0\%
				& --
				& 38.7\%$\pm$6.1\% \\
				
				& \cellcolor{gray!15}\textbf{+FMC}
				& \cellcolor{gray!15}89.3\%$\pm$2.3\%
				& \cellcolor{gray!15} 37.3\%$\pm$6.1\%
				& \cellcolor{gray!15}52.0\%$\pm$10.6\%
				& \cellcolor{gray!15}69.3\%$\pm$8.3\%
				& \cellcolor{gray!15}0.019$\pm$0.002
				& \cellcolor{gray!15}69.3\%$\pm$8.3\% \\
				\cline{1-8}
				
				\multirow{2}{*}{HC}
				& +AIC
				& 48.0\%$\pm$0.0\%
				& 33.3\%$\pm$4.6\%
				& 37.3\%$\pm$8.3\%
				& 42.7\%$\pm$6.1\%
				& --
				& 42.7\%$\pm$6.1\% \\
				
				& \cellcolor{gray!15}\textbf{+FMC}
				& \cellcolor{gray!15}\textbf{92.0\%$\pm$0.0\%}
				& \cellcolor{gray!15} 42.7\%$\pm$8.3\%
				& \cellcolor{gray!15}60.0\%$\pm$4.0\%
				& \cellcolor{gray!15}72.0\%$\pm$4.0\%
				& \cellcolor{gray!15}0.034$\pm$0.023
				& \cellcolor{gray!15}56.0\%$\pm$8.0\% \\
				\cline{1-8}
				
				\multirow{2}{*}{BP}
				& +AIC
				& 48.0\%$\pm$0.0\%
				& 32.0\%$\pm$0.0\%
				& 48.0\%$\pm$0.0\%
				& 48.0\%$\pm$0.0\%
				& --
				& 48.0\%$\pm$0.0\% \\
				
				& \cellcolor{gray!15}\textbf{+FMC}
				& \cellcolor{gray!15}\textbf{92.0\%$\pm$0.0\%}
				& \cellcolor{gray!15} \textbf{48.0\%$\pm$0.0\%}
				& \cellcolor{gray!15}\textbf{72.0\%$\pm$0.0\%}
				& \cellcolor{gray!15}\textbf{84.0\%$\pm$0.0\%}
				& \cellcolor{gray!15}\textbf{0.009$\pm$0.007}
				& \cellcolor{gray!15}\textbf{84.0\%$\pm$0.0\%} \\
				
				\midrule[.1em]
				\multicolumn{8}{c}{\textbf{Thor}} \\
				\midrule
				\multicolumn{2}{c!{\vrule width 1.2pt}}{\textbf{Method}}
				& \textbf{Answered ($\uparrow$)}
				& \textbf{Feas.@0\% ($\uparrow$)}
				& \textbf{Feas.@5\% ($\uparrow$)}
				& \textbf{Feas.@10\% ($\uparrow$)}
				& \textbf{Gap$\cap$@10\% ($\downarrow$)}
				& \textbf{$\geq$uniform@10\% ($\uparrow$)} \\
				\midrule
				
				\multirow{2}{*}{Rand}
				& +AIC
				& 16.0\%$\pm$0.0\%
				& 9.3\%$\pm$2.3\%.
				& 10.7\%$\pm$4.6\%
				& 13.3\%$\pm$2.3\%
				& --
				& 4.0\%$\pm$6.9\% \\
				
				& \cellcolor{gray!15}\textbf{+FMC}
				& \cellcolor{gray!15}72.0\%$\pm$0.0\%
				& \cellcolor{gray!15} 34.7\%$\pm$2.3\%
				& \cellcolor{gray!15}49.3\%$\pm$2.3\%
				& \cellcolor{gray!15}62.7\%$\pm$2.3\%
				& \cellcolor{gray!15}0.007$\pm$0.002
				& \cellcolor{gray!15}37.3\%$\pm$2.3\% \\
				\cline{1-8}
				
				\multirow{2}{*}{HC}
				& +AIC
				& 16.0\%$\pm$0.0\%
				& 8.0\%$\pm$0.0\%
				& 10.7\%$\pm$2.3\%
				& 12.0\%$\pm$0.0\%
				& --
				& 1.3\%$\pm$2.3\% \\
				
				& \cellcolor{gray!15}\textbf{+FMC}
				& \cellcolor{gray!15}72.0\%$\pm$0.0\%
				& \cellcolor{gray!15} 37.3\%$\pm$4.6\%
				& \cellcolor{gray!15}50.7\%$\pm$6.1\%
				& \cellcolor{gray!15}56.0\%$\pm$4.0\%
				& \cellcolor{gray!15}0.008$\pm$0.001
				& \cellcolor{gray!15}29.3\%$\pm$4.6\% \\
				\cline{1-8}
				
				\multirow{2}{*}{BP}
				& +AIC
				& 16.0\%$\pm$0.0\%
				& 16.0\%$\pm$0.0\%
				& 16.0\%$\pm$0.0\%
				& 16.0\%$\pm$0.0\%
				& --
				& 16.0\%$\pm$0.0\% \\
				
				& \cellcolor{gray!15}\textbf{+FMC}
				& \cellcolor{gray!15}\textbf{76.0\%$\pm$0.0\%}
				& \cellcolor{gray!15}\textbf{40.0\%$\pm$0.0\%}
				& \cellcolor{gray!15}\textbf{56.0\%$\pm$0.0\%}
				& \cellcolor{gray!15}\textbf{64.0\%$\pm$0.0\%}
				& \cellcolor{gray!15}\textbf{0.002$\pm$0.001}
				& \cellcolor{gray!15}\textbf{60.0\%$\pm$0.0\%} \\
				\bottomrule[.1em]
			\end{tabular}
		\end{adjustbox}
	\end{table}

	
	First, we compare the answer rates and deployment feasibility of search algorithms
	using vanilla AIConfigurator (AIC) and our FMC. Table~\ref{tab:aic_vs_FMC} shows that
	FMC consistently improves both answer rate and measured deployment feasibility
	across all search algorithms. On A6000, FMC raises answer rates from 48.0\% to
	at least 89.3\%, while on Thor it increases them from 16.0\% to at least 72.0\%.
	The improvement is also reflected in measured feasibility at all three tolerance
	levels. For BP, strict feasibility increases from 32.0\% to 48.0\% on A6000 and
	from 16.0\% to 40.0\% on Thor; with a 10\% tolerance, BP+FMC reaches 84.0\%
	and 64.0\%, respectively. Thus, FMC substantially improves the usefulness of AIC
	predictions for configuration admission, although prediction errors near the SLO
	boundary can still lead to violations under strict feasibility.
	
	Among the FMC-based methods, BP achieves the highest feasibility@10\% and the
	smallest quality gap on both platforms. On the subset of requests for which Rand,
	HC, and BP are all feasible@10\%, BP has an average RMQD gap of
	$0.009\pm0.007$ on A6000 and $0.002\pm0.001$ on Thor relative to the best
	measured feasible@10\% configuration in the evaluation dataset. BP also matches
	or improves upon the best measured feasible@10\% uniform configuration on
	84.0\% and 60.0\% of all requests, respectively. These results indicate that
	combining calibrated latency prediction with structured mixed-precision search
	provides a favorable quality-feasibility trade-off under the evaluated deployment
	requests.
	
	
	\begin{table}[h]
		\centering
		\caption{Latency prediction error as median absolute percentage error (MdAPE) on the collected mixed-precision (layerwise) LLM evaluation dataset. AIC~\citep{xu2026aiconfigurator} does not require end-to-end latency data. For LR~\citep{imai2024predicting}, FMC-MSE, and FMC, we use end-to-end latency data of 30 configurations evenly split across backends.}
		\label{tab:latency_prediction}
		\begin{adjustbox}{max width=\linewidth}	
			\begin{tabular}{c|c|ccc|c|ccc|c}
				\toprule[.1em]
				\multirow{2}{*}{\textbf{Platform}}
				& \multirow{2}{*}{\textbf{Backend}}
				& \multicolumn{4}{c|}{\textbf{TTFT MdAPE\% ($\downarrow$)}}
				& \multicolumn{4}{c}{\textbf{TPOT MdAPE\% ($\downarrow$)}} \\
				\cline{3-10}
				
				& &  AIC &  LR &  FMC-MSE
				& FMC \textbf{(Ours)}
				&  AIC &  LR &  FMC-MSE
				&FMC \textbf{(Ours)} \\
				\hline
				\multirow{3}{*}{A6000} & vLLM & 78.5 & 55.4 &  43.3 & \textbf{24.8} &11.4 & 100.0 &76.9 & {\bf 5.8}\\ 
				& SGLang & 65.2 & 40.6 & 44.3 & \textbf{34.3} & \textbf{5.8} & 100.0 &  132.5 &  8.8\\ 
				& TRT-LLM & 216.9 & 100.0 & 40.8 &  {\bf 30.9} & {\bf 17.7} & 347.4 & 633.3  & 19.7\\ 
				\hline
				\multirow{2}{*}{Thor}& vLLM & 141.5 & 100.0 & 128.6 & \textbf{30.0} &   39.9 & 100.0 & 287.0  & \textbf{20.6}\\ 
				& SGLang & 62.8 & 175.7 & 75.4    &\textbf{33.1} & 10.7 &344.8  &  81.3  & {\bf 9.3}\\ 
				\bottomrule
		\end{tabular}\end{adjustbox}
	\end{table}
	
	\subsection{Ablation Study}\label{sec:ablation}
	To justify our choice of latency surrogate, we compare FMC with vanilla AIC and
	two baselines on the collected evaluation data: (1) a linear regression model that
	predicts end-to-end latency from configuration features and a roofline-based
	analytical estimate (LR)~\citep{imai2024predicting}, and (2) FMC trained with
	mean-squared error (FMC-MSE) instead of the pinball loss in Eq.~\eqref{eq:pinball}.
	As shown in Table~\ref{tab:latency_prediction}, FMC achieves the lowest TTFT
	prediction error for every platform-backend pair, reducing AIC's average TTFT
	MdAPE from 113.0\% to 30.6\%. It also outperforms LR and FMC-MSE in all TTFT
	settings. For TPOT, FMC achieves the lowest error in three of the five settings
	and remains close to AIC in the other two. In these cases, the raw AIC estimates
	are already relatively accurate, leaving limited room for improvement, and
	few-shot calibration may overcorrect them (We notice that the fitted slopes are close to one). In contrast, LR and FMC-MSE
	frequently exhibit errors above 100\%, particularly for TPOT, suggesting that
	direct regression and squared-loss calibration are unstable when only a few
	profiling samples are available. These results show that robust log-space median calibration substantially reduces
	the latency-prediction error of AIC with few end-to-end profiles.
	
	
	We also evaluate the quality surrogate alone
	by comparing the constrained ridge regression in Section~3.2 with
	unconstrained ridge regression using the same features. Both surrogates are trained on the 226 configurations
	described in Section~4.2 and evaluated on the held-out configurations in the evaluation data. In Table~\ref{tab:quality-surrogate}, we report median absolute percentage error (MdAPE) between predicted
	and measured RMQD, which shows 
	the effect of the monotonicity constraints. 
	\begin{table}[t]
		\centering
		\caption{Quality-surrogate RMQD prediction error on held-out
			configurations. Lower MdAPE is better.}
		\label{tab:quality-surrogate}
		\begin{tabular}{lcc}
			\hline
			& Unconstrained ridge & Constrained ridge (Ours) \\
			\hline
			MdAPE (\%) $\downarrow$ & 3.3 & \textbf{2.6} \\
			\hline
		\end{tabular}
	\end{table}


	\section{Conclusion}
	
	We presented \algname, a surrogate-guided framework for selecting LLM inference configurations with highest predicted generation quality under hardware memory and request-specific latency constraints. \algname combines RMQD, a computationally efficient proxy for generation quality; a lightweight quality surrogate that predicts RMQD directly from model configurations; FMC, which calibrates analytical latency estimates using a small number of end-to-end profiles; and efficient search algorithms for exploring combinatorial layerwise mixed-precision spaces. Across deployment requests spanning model families, serving backends, and input/output sequence lengths, our experiments show that calibrated latency gates and structured mixed-precision search identify more hardware-feasible, high-quality configurations than uncalibrated/unstructured  baselines. Future work may explore agentic LLMs as surrogates or search algorithms for navigating increasingly complex deployment spaces.

	\bibliographystyle{tmlr}
	\bibliography{draft}
	
	
	\appendix
	
	
	
	\section{Ablation Study of $\eta$ in BandPareto}\label{sec:ablation-eta}
	
	On A6000, we sweep BandPareto's soft-screening slack $\eta$ over
	$\{0\%, 5\%, 10\%, 15\%, 25\%\}$. As shown in
	Table~\ref{tab:eta-ablation}, increasing $\eta$ from $0\%$ to $5\%$
	improves feasibility ratio. Increasing $\eta$ beyond $5\%$
	provides no further feasibility gains on these requests, while
	increasing the number of evaluated candidates and wall-clock time. 
	
	\begin{table}[h]
		\centering
		\caption{Measured deployment feasibility, number of evaluated candidates after FracMix, and wall-clock time (per request) of BandPareto with $\eta = 0\%, 5\%, 10\%, 15\%, 25\%$.}
		\label{tab:eta-ablation}
		\begin{tabular}{ccccc}
			\toprule
			$\eta$ & Feas.@$5\%$ ($\uparrow$) & Feas.@$10\%$ ($\uparrow$) & \# Candidates & Wall-Clock Time (s) \\
			\midrule
			$0\%$ & 68\% & 80\% & 24 & 2.2 \\
			$5\%$  & 72\%  & 84\% & 43 & 3.5 \\
			$10\%$ & 72\%  & 84\% & 61 & 4.5 \\
			$15\%$  & 72\%  & 84\% & 76 & 5.6 \\
			$25\%$  & 72\%  & 84\% & 96 & 7.2 \\
			\bottomrule
		\end{tabular}
	\end{table}
	
	\section{List of Deployment requests for Search Performance Comparison}\label{sec:requests}
	
	To compare search performance, we generate 50 synthetic deployment requests using a fixed procedure. Each request specifies a platform–backend pair and an $(\mathrm{isl},\mathrm{osl})$ workload, where $isl \in \{128, 512, 2048, 8192\}$ and $osl \in \{64, 256, 512\}$. We derive $L_1$ and $L_2$ by selecting an anchor model from
	\{Llama-3.2-3B, Qwen3-4B, Qwen3-8B\} and a batch size from
	$\{4,8,32\}$, setting the latency constraint to be at least 5\% below that of the
	corresponding uniform \texttt{int8} configuration (on A6000) or
	\texttt{fp8} configuration (on Thor). Full lists of requests are in Table~\ref{tab:a6000-exp2-requests} and Table~\ref{tab:thor-requests}.
	
	\begin{table}[h]
		\centering
		\small
		\caption{Requests on A6000.}
		\label{tab:a6000-exp2-requests}
		\setlength{\tabcolsep}{3.5pt}
		\begin{tabular}{r l l r r r r}
			\toprule
			\# & Backend & Model Family & ISL & OSL & $L_1$ & $L_2$ \\
			\midrule
			0 & sglang & \textsc{Llama-3.2}  & 8192 & 64 & 276.8 & 19.60 \\
			1 & sglang & \textsc{Llama-3.2}  & 8192 & 256 & 303.9 & 15.00 \\
			2 & vllm & \textsc{Qwen3}  & 2048 & 64 & 577.5 & 35.63 \\
			3 & sglang & \textsc{Qwen3}  & 8192 & 512 & 1066.1 & 38.73 \\
			4 & vllm & \textsc{Qwen3}  & 512 & 64 & 157.6 & 31.25 \\
			5 & sglang & \textsc{Qwen3}  & 8192 & 256 & 1036.7 & 29.38 \\
			6 & vllm & \textsc{Qwen3}  & 2048 & 64 & 279.7 & 31.81 \\
			7 & vllm & \textsc{Qwen3}  & 8192 & 64 & 964.6 & 56.95 \\
			8 & vllm & \textsc{Qwen3}  & 8192 & 512 & 1769.1 & 54.26 \\
			9 & sglang & \textsc{Qwen3}  & 2048 & 64 & 507.9 & 18.66 \\
			10 & sglang & \textsc{Qwen3}  & 2048 & 256 & 449.7 & 18.62 \\
			11 & sglang & \textsc{Qwen3}  & 8192 & 512 & 1040.9 & 22.35 \\
			12 & vllm & \textsc{Qwen3}  & 2048 & 64 & 293.2 & 18.40 \\
			13 & sglang & \textsc{Qwen3}  & 2048 & 512 & 450.9 & 15.67 \\
			14 & sglang & \textsc{Qwen3}  & 2048 & 256 & 482.5 & 12.41 \\
			15 & sglang & \textsc{Qwen3}  & 2048 & 512 & 449.0 & 11.48 \\
			16 & vllm & \textsc{Qwen3}  & 2048 & 512 & 268.4 & 14.70 \\
			17 & vllm & \textsc{Qwen3}  & 2048 & 256 & 274.8 & 12.14 \\
			18 & sglang & \textsc{Qwen3}  & 512 & 256 & 219.1 & 9.12 \\
			19 & sglang & \textsc{Qwen3}  & 512 & 256 & 415.3 & 9.91 \\
			20 & sglang & \textsc{Qwen3}  & 128 & 512 & 117.5 & 9.49 \\
			21 & sglang & \textsc{Qwen3}  & 512 & 512 & 419.0 & 10.11 \\
			22 & sglang & \textsc{Qwen3}  & 128 & 256 & 117.1 & 9.27 \\
			23 & sglang & \textsc{Qwen3}  & 512 & 64 & 227.9 & 8.97 \\
			24 & sglang & \textsc{Qwen3}  & 512 & 512 & 226.2 & 9.22 \\
			\bottomrule
		\end{tabular}
	\end{table}
	
	\begin{table}[h]
		\centering
		\small
		\caption{Requests on Thor.}
		\label{tab:thor-requests}
		\setlength{\tabcolsep}{3.5pt}
		\begin{tabular}{r l l r r r r}
			\toprule
			\# & Backend & Model Family & ISL & OSL & $L_1$ & $L_2$ \\
			\midrule
			0 & sglang & \textsc{Llama-3.2}  & 8192 & 64 & 888.4 & 111.99 \\
			1 & vllm & \textsc{Llama-3.2}   & 8192 & 64 & 692.5 & 456.97 \\
			2 & vllm & \textsc{Llama-3.2}  & 8192 & 64 & 699.5 & 110.17 \\
			3 & sglang & \textsc{Llama-3.2}   & 2048 & 64 & 672.9 & 99.66 \\
			4 & vllm & \textsc{Llama-3.2}   & 8192 & 512 & 700.7 & 175.55 \\
			5 & vllm & \textsc{Qwen3}  & 8192 & 64 & 1030.0 & 150.84 \\
			6 & sglang & \textsc{Qwen3}  & 8192 & 64 & 1228.3 & 155.38 \\
			7 & sglang & \textsc{Qwen3}  & 8192 & 64 & 1777.2 & 859.79 \\
			8 & sglang & \textsc{Qwen3}  & 8192 & 64 & 1617.5 & 217.47 \\
			9 & vllm & \textsc{Qwen3}  & 8192 & 64 & 1428.9 & 199.92 \\
			10 & vllm & \textsc{Qwen3} & 8192 & 64 & 1028.1 & 646.13 \\
			11 & sglang & \textsc{Qwen3}  & 2048 & 64 & 855.3 & 125.22 \\
			12 & vllm & \textsc{Qwen3} & 2048 & 64 & 147.3 & 36.62 \\
			13 & vllm & \textsc{Qwen3}  & 2048 & 512 & 455.1 & 65.67 \\
			14 & vllm & \textsc{Qwen3}  & 8192 & 512 & 1033.6 & 65.28 \\
			15 & vllm & \textsc{Qwen3}  & 2048 & 512 & 291.7 & 26.81 \\
			16 & sglang & \textsc{Qwen3}   & 128 & 512 & 332.7 & 27.67 \\
			17 & sglang & \textsc{Qwen3}  & 2048 & 64 & 1211.7 & 55.85 \\
			18 & sglang & \textsc{Qwen3}   & 8192 & 512 & 1631.7 & 89.72 \\
			19 & sglang & \textsc{Qwen3}  & 2048 & 512 & 1212.5 & 43.30 \\
			20 & sglang & \textsc{Qwen3}  & 128 & 512 & 146.0 & 31.73 \\
			21 & vllm & \textsc{Qwen3}  & 128 & 512 & 138.1 & 29.95 \\
			22 & vllm & \textsc{Qwen3}  & 128 & 64 & 87.0 & 22.21 \\
			23 & sglang & \textsc{Qwen3}   & 128 & 512 & 92.3 & 18.20 \\
			24 & vllm & \textsc{Qwen3}  & 128 & 512 & 138.9 & 16.41 \\
			\bottomrule
		\end{tabular}
	\end{table}
	
	
	

\end{document}